\pdfoutput=1

\documentclass[11pt]{article}

\usepackage[final]{acl}

\usepackage{times}
\usepackage{latexsym}

\usepackage[T1]{fontenc}

\usepackage[utf8]{inputenc}

\usepackage{microtype}

\usepackage{inconsolata}

\usepackage{graphicx}

\usepackage{amsmath}
\usepackage{amssymb}
\usepackage{mathtools}
\usepackage{amsthm}
\usepackage{pifont}
\usepackage{subcaption}

\usepackage{multirow}
\usepackage{booktabs}

\definecolor{forestgreen}{rgb}{0.0, 0.5, 0.0}

\usepackage{bbm}
\usepackage{hyperref}
\usepackage{enumitem}

\newcommand{\mydeltax}[1]{\colorbox[RGB]{243, 243, 243}{\makebox(85,6){#1}}}
\newcommand{\mydelta}[1]{\colorbox[RGB]{243, 243, 243}{\makebox(32,6){#1}}}
\newcommand{\dataset}[0]{MVP-Bench}
\definecolor{forestgreen}{rgb}{0.0, 0.5, 0.0}
\usepackage{xcolor, soul,todonotes} 
	\definecolor{kmycolor}{rgb}{0.858, 0.188, 0.478}

\title{MVP-Bench: Can Large Vision--Language Models Conduct Multi-level Visual Perception Like Humans?}

\author{Guanzhen Li \quad Yuxi Xie \thanks{Corresponding author} \quad Min-Yen Kan \\
National University of Singapore \\
\texttt{\{guanzhen,xieyuxi\}@u.nus.edu} \quad \texttt{kanmy@comp.nus.edu.sg}
}

\begin{document}
\maketitle
\begin{abstract}
Humans perform visual perception at multiple levels, including low-level object recognition and high-level semantic interpretation such as behavior understanding.
Subtle differences in low-level details can lead to substantial changes in high-level perception.
For example, substituting the shopping bag held by a person with a gun suggests violent behavior, implying criminal or violent activity.
Despite significant advancements in various multimodal tasks, Large Visual Language Models (LVLMs) remain unexplored in their capabilities to conduct such multi-level visual perceptions.

To investigate the perception gap between LVLMs and humans, we introduce \dataset{}, the first visual--language benchmark systematically evaluating both low- and high-level visual perception of LVLMs.
We construct \dataset{} across natural and synthetic images to investigate how manipulated content influences model perception. 
Using MVP-Bench, we diagnose the visual perception of 10 open-source and 2 closed-source LVLMs, showing that high-level perception tasks significantly challenge existing LVLMs. The state-of-the-art GPT-4o only achieves an accuracy of $56\%$ on Yes/No questions, compared with $74\%$ in low-level scenarios. Furthermore, the performance gap between natural and manipulated images indicates that current LVLMs do not generalize in understanding the visual semantics of synthetic images as humans do.
Our data and code are publicly available at \href{https://github.com/GuanzhenLi/MVP-Bench}{https://github.com/GuanzhenLi/MVP-Bench}.
\end{abstract}

\section{Introduction}
Visual perception (VP) refers to the ability to transform visual signals into meaningful perceptions \cite{DEWIT2012665,gordon2019intermodulation}. When humans parse visual signals, they initially engage in high-level perception to grasp the overarching concept using commonsense knowledge. This serves as context guidance for exploring further low-level details aligned with their intentions \cite{wang2024browse,garner1987metacognition}. For example, given an image of a man in a bar, humans first grasp the high-level concept, such as the behaviour of drinking, and focus on low-level details, such as the type of alcohol, to obtain specific information. Existing Large Vision--Language Models (LVLMs) demonstrate an exceptional understanding of such low-level visual clues. However, it remains unexplored whether they have similar hierarchical visual perceptions at both levels, like humans.


\begin{figure*}[htbp]
    \centering
    \includegraphics[width=\textwidth]{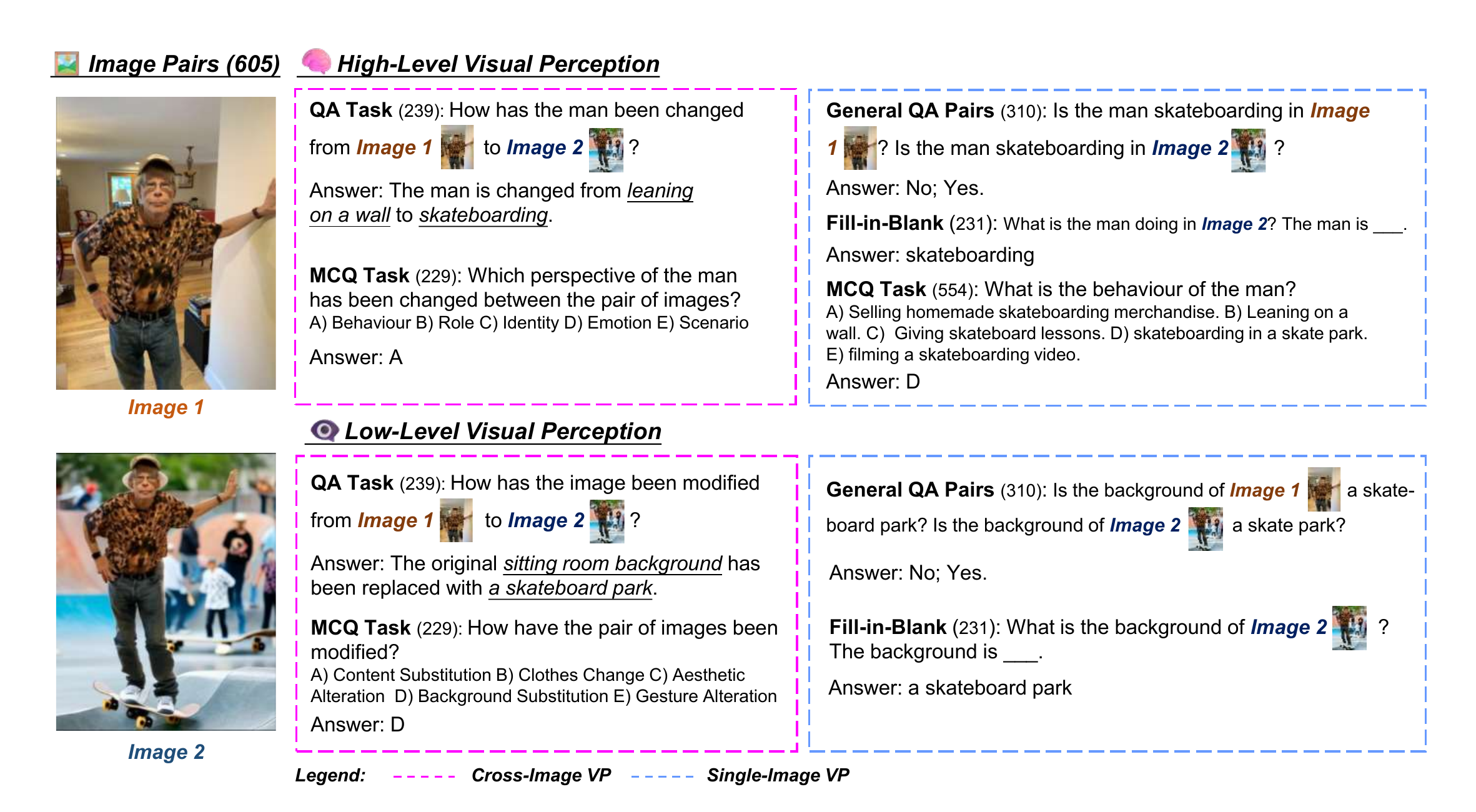}
    \caption{A sample of \dataset{} manifesting both high- and low-level visual perception. \textit{Image 1} and \textit{Image 2} form an image pair. Their different backgrounds indicate that the man is engaged in different behaviours.}
    \label{fig1}
    \vskip -0.1in
\end{figure*} 
Recently, several benchmarking works have considered evaluating visual perceptions~\cite{liu2023mmbench,fu2024mme,chow2021travlr}. However, such holistic evaluation benchmarks lack the critical specialization needed to assess visual perceptions. Specifically, most of their tasks focus on low-level perception such as \textit{Counting} and \textit{Existence Detection} questions on single images. Besides, existing benchmarks are mostly designed based on individual question--image samples, failing to evaluate the consistency and accuracy of understanding an image with different forms of perceptions. Furthermore, most of the current benchmarks are built on real-world natural image data, making it hard to disentangle reliance on prior knowledge from the visual perception of specific contexts, such as synthetic images~\cite{bitton2023breaking}. Motivated by the challenges of interpreting LVLMs' visual perception capabilities, we propose \dataset{}, the first benchmark systematically evaluating multi-level visual perceptions of LVLMs. As shown in Figure~\ref{fig1}, each sample is accompanied by questions at both levels. We thoroughly design five high-level and thirteen low-level perception categories, detailed in Section~\ref{3}. Furthermore, we construct \{natural, manipulated\} image pairs which convey contrasting perceptions as a more challenging task for visual perception. 

In this work, with our constructed \dataset{}, we evaluate twelve LVLMs and find that there is a significant performance gap between high- and low-level visual perception in LVLMs. Furthermore,  we observe that manipulated visual contents are more challenging than natural images for LVLMs to understand and interpret. Our further qualitative analysis reveals the deficiency of current LVLMs and the gap between open- and closed-source models.
\section{Related Work}
\paragraph{Visual Perception.} 
Visual Perception represents how the human brain transforms the pattern of information on the retina into a meaningful perception of the world \cite{DEWIT2012665,cornsweet2012visual}. 
This process involves interactions among sensory and cognitive processes across hierarchical levels in the brain \cite{gordon2019intermodulation,ROUW1997209}. 
Low-level visual features refer to the properties like colors and spatial attributes, while high-level visual processing integrates with human cognitive functions (\textit{e.g.} commonsense knowledge, personal experiences) related to  recognized objects \cite{akcelik2022influence,wu2023q,1180642113,SCHINDLER202114}. 
Both perception competences are crucial, as human visual perception begins with grasping the image's main idea at a high level, and then delving into low-level features motivated by particular intentions \cite{garner1987metacognition}. 
In \dataset{}, we define five high-level categories and thirteen low-level categories. The mapping relationships between levels indicate that certain low-level features can support the high-level perception (illustrated in Section~\ref{3}).

\paragraph{Vision--Language Benchmarks.} 
Some recent benchmarks contain visual perception as a section, but their aim to offer a comprehensive evaluation of LVLMs' various capabilities leads to an inadequate exploration of visual perception. MMBench \cite{liu2023mmbench} and MME \cite{fu2024mme} categorize visual perception based on question granularity. Although coarse perception questions are general, their questions like \textit{Counting} or \textit{Existence Detection} cannot reflect an image's main idea as high-level visual perception. Additionally, they evaluate different categories of visual perception individually, making it unavailable to compare an LVLM's different perceptions.
II-Bench\cite{liu2024ii} and DEEPEVAL\cite{yang2024can} focus on understanding deep image semantics, requiring LVLMs to perform complex commonsense reasoning based on low-level details. While they also reveal LVLMs' performance gaps across levels, these gaps are primarily due to limited reasoning abilities rather than intuitive visual perception.
The definition of perception in PCA-Bench \cite{chen2024pca} resembles our benchmark, emphasizing how perception offers a guiding context in decision-making domains. However, their images depicting environments normally do not require significant high-level perception.
\dataset{} systematically evaluates LVLMs' multi-level visual perception, with each image accompanied by high- and low-level questions simultaneously. As perceptions related to humans normally require significant perception at both levels (such as misinformation understanding or emotion recognition) \cite{peng2023agenda,thomson2022visual}, we construct image pairs containing humans to ensure that the cases can assess LVLMs' multi-level perception.

\paragraph{Synthetic Images.} 
Recent advancements in image generation tools \cite{ramesh2021zero,rombach2021highresolution} and image editing models \cite{brooks2023instructpix2pix,zhang2023sine} have led to synthetic datasets for different tasks, such as Whoops \cite{bitton2023breaking} and StableRep \cite{tian2024stablerep}. In the process of utilizing text-to-image tools for generating synthetic images, a prompt aligned with the expected image content is essential. In previous works, the source of such prompts can be manually-crafted prompts \cite{bitton2023breaking}, text annotations in existing datasets \cite{tian2024stablerep} or prompts generated by LLMs \cite{aboutalebi2024magid,li2023stablellava,wu2023visual}. 
In \dataset{}, we generate manipulated images for constructing image pairs. To obtain a prompt tailored to each case while minimizing human effort, we employ ChatGPT to generate the prompts (\textit{cf.} Section~\ref{4.1}).
\section{MVP-Bench Evaluation Suite}
\label{3}
MVP-Bench comprises $530$ \{natural, manipulated\} image pairs accompanied by questions at multiple perception levels. Using MVP-Bench, we diagnose LVLMs by investigating (1) the performance gap between high- and low-level visual perceptions and (2) the difference in visual understanding abilities on natural and manipulated images.

\subsection{Evaluation across Perception Levels}
We prioritize the perception of humans as high-level perception, \textit{e.g.}, misinformation understanding \cite{da2021edited} and emotion recognition \cite{hari2009brain}, where high-level perception is commonly engaged. 

We categorize high-level ({$L_h$}) perceptions of humans into five dimensions, including \textit{Behaviour}, \textit{Role}, \textit{Identity}, \textit{Emotion}, \textit{Scenario}. 
Each dimension corresponds to several low-level (${L_l}$) perception types. As shown in Figure~\ref{stat} (a), certain low-level perceptions (\textit{e.g.}, \textit{attire} such as a police uniform or \textit{group association} with firefighters) can support the high-level perception (\textit{e.g.}, \textit{Role}).

We design Yes/No questions and Cross-Image questions at both levels. Constructed on the same set of images, the multi-level perception tasks enable us to diagnose the perception gap in LVLMs across different levels. Specifically, we calculate the accuracy on Yes/No questions based on the correctness of each individual question--image pair (represented as $aAcc$), while all multiple-choice questions within \dataset{} are evaluated with Circular Strategy \cite{liu2023mmbench} to alleviate the model prediction bias from the option order.

\subsection{Evaluation with Image Pairs}

Each \{natural, manipulated\} image pair in \dataset{} conveys significantly different multi-level perceptions. Specifically, the two images differ only in one of the ${L_l}$ perception categories (in Figure~\ref{stat} (a)), leading to distinct ${L_h}$ perceptions. 
To mitigate the effect of the LVLMs' biased tendency to answer Yes/No questions~\cite{liu2023hallusionbench}, we examine if LVLMs can elicit different perceptions given an image pair with the same question. We further explore the performance gap in LVLMs on natural and manipulated images in Section~\ref{5}.

For Yes/No questions, we ask the same question on pairwise image data. As the two images are manipulated to convey different perceptions, they have opposite corresponding ground truth answers. We calculate $qAcc$ and $iAcc$ based on question- and image-level accuracy, respectively, following~\citep{liu2023hallusionbench}. We design a holistic metric $mAcc$, requiring answering all questions corresponding to an image pair correctly.

For single-image multiple-choice questions, we focus on model understanding of manipulated images as a more challenging task. We include the answer to the natural image as a distractor to assess the discriminability of LVLMs in discerning the differences between the image pair. Additionally, we leverage ChatGPT\footnote{We used \texttt{gpt-3.5-turbo-1106}.} to generate three other options aligned with the low-level clues in the manipulated image to heighten our task difficulty.

\definecolor{f2_blue}{rgb}{0.2656,0.4453,0.7656}
\definecolor{f2_orange}{rgb}{0.9258,0.4883,0.1914}
\definecolor{f2_green}{rgb}{0.3281,0.5078,0.2070}
\definecolor{f2_purple}{rgb}{0.5430,0.3438,0.6914}

\begin{figure*}[htbp]
    \centering
    \includegraphics[width=.9\textwidth]{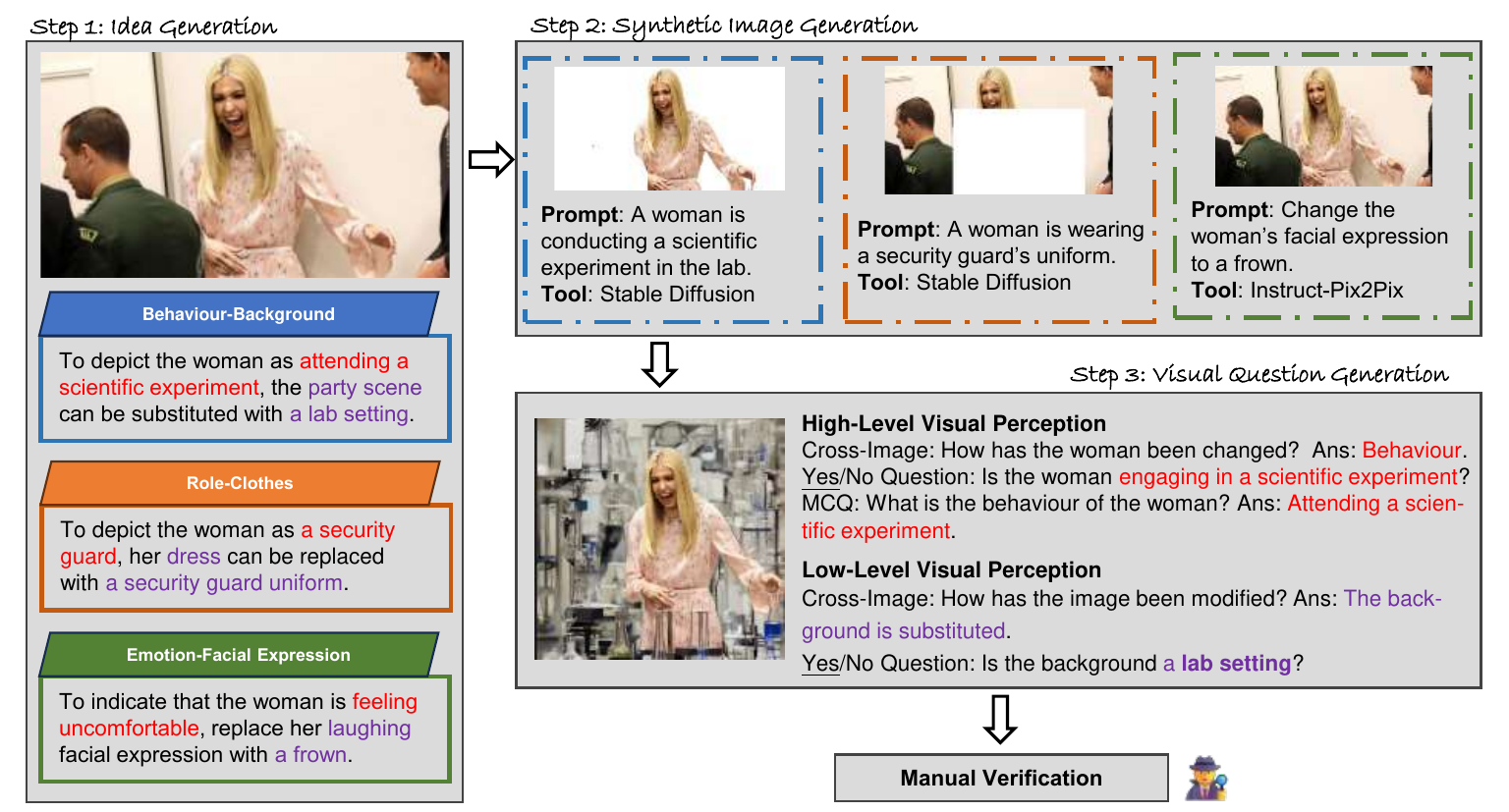}
    \caption{\dataset{} three-step construction pipeline (best viewed in color). Step 1 uses three categories (\textcolor{f2_blue}{`Behaviour-Background'}, \textcolor{f2_orange}{`Role-Clothes'}, \textcolor{f2_green}{`Emotion-Facial Expression'}) as examples to illustrate how \textcolor{red}{high-level perception} guides the identification of \textcolor{f2_purple}{low-level perception}. Step 2 demonstrates three categories of manipulated image generation: \textit{Overall Background Substitution}, \textit{Partial Component Substitution}, and \textit{Direct Alteration} (from left to right). Step 3 explains how to generate questions based on the ideas obtained in Step 1, with the same colour indicating that the generated question is based on the corresponding part from the expected perception.}
    \label{fig2}
    \vskip -0.1in
\end{figure*} 

\section{MVP-Bench Construction}

We now present our construction process of image manipulation and the designs of corresponding multi-level questions for \dataset{}. 

\subsection{Construction Pipeline}
\label{4.1}
We select images from the EMU dataset \cite{da2021edited} as natural images for constructing image pairs. EMU focuses on visual misinformation, portraying cases involving humans and complex social scenes that require perceptions at both levels. Based on the natural image, we generate synthetic manipulations following one of the ${L_l}$ categories.

However, to alter manipulated images' ${L_h}$ perceptions in certain categories, it is challenging to constrain the manipulation applied exactly to a specific ${L_l}$ category without significant modification on other details. Besides, it is also hard to ensure consistency between the image pairs and the questions. We propose a three-step benchmark construction pipeline to meet the two requirements.

\paragraph{Step one: Idea Generation.} We utilize ChatGPT to generate ideas on how to manipulate natural images via Chain of Thoughts (CoT). Given an initially determined ${L_h}$ category, we prompt ChatGPT to identify a corresponding low-level perception to support it. For instance, in Figure~\ref{fig2}, considering the ``Behaviour-Background Substitution'' category, ChatGPT first generates an idea to change the woman's behaviour from attending a party to engaging in an experiment. Under this guidance, the background of the manipulated image should be a laboratory environment. Specifically, we provide auxiliary information such as the description of the manipulated image, which is incorporated into the textual prompt for image generation in Step 2.

To ensure coherence between the generated idea and the subsequent visual editing, we fixate on a specific subject at this initial step utilizing the visual grounding ability of Shikra~\citep{chen2023shikra}. Specifically, we employ Shikra to retrieve the coordinates of a selected subject ($C_{sub}$) and utilize it to query low-level features (\textit{e.g.}, ``What is the man holding?'') from the image in the subsequent steps.

\paragraph{Step two: Manipulated Image Generation.} We define three categories of manipulated image generation based on the image-editing type: Partial Component Substitution, Overall Background Substitution, and Direct Manipulation.

\textit{2.1 Partial Component Substitution.} This refers to manipulating an image by substituting an object or a part of the main subject. The pipeline utilizes Shikra to extract the target object's coordinates ($C_{obj}$), with $C_{sub}$ serving as a constraint. After masking $C_{obj}$ as a blank, we apply the Stable-Diffusion-Inpaint \cite{stacchio2023stableinpainting} as a tool, using the edited image's caption obtained from step one as the prompt to generate a manipulated image. A set of defined ${L_l}$ categories, $\{B_2,B_3,B_4,R_2,I_1,I_2,I_3,E_1\}$, can be executed in this process.

\begin{figure*}[h]
    \centering
    \includegraphics[width=\textwidth]{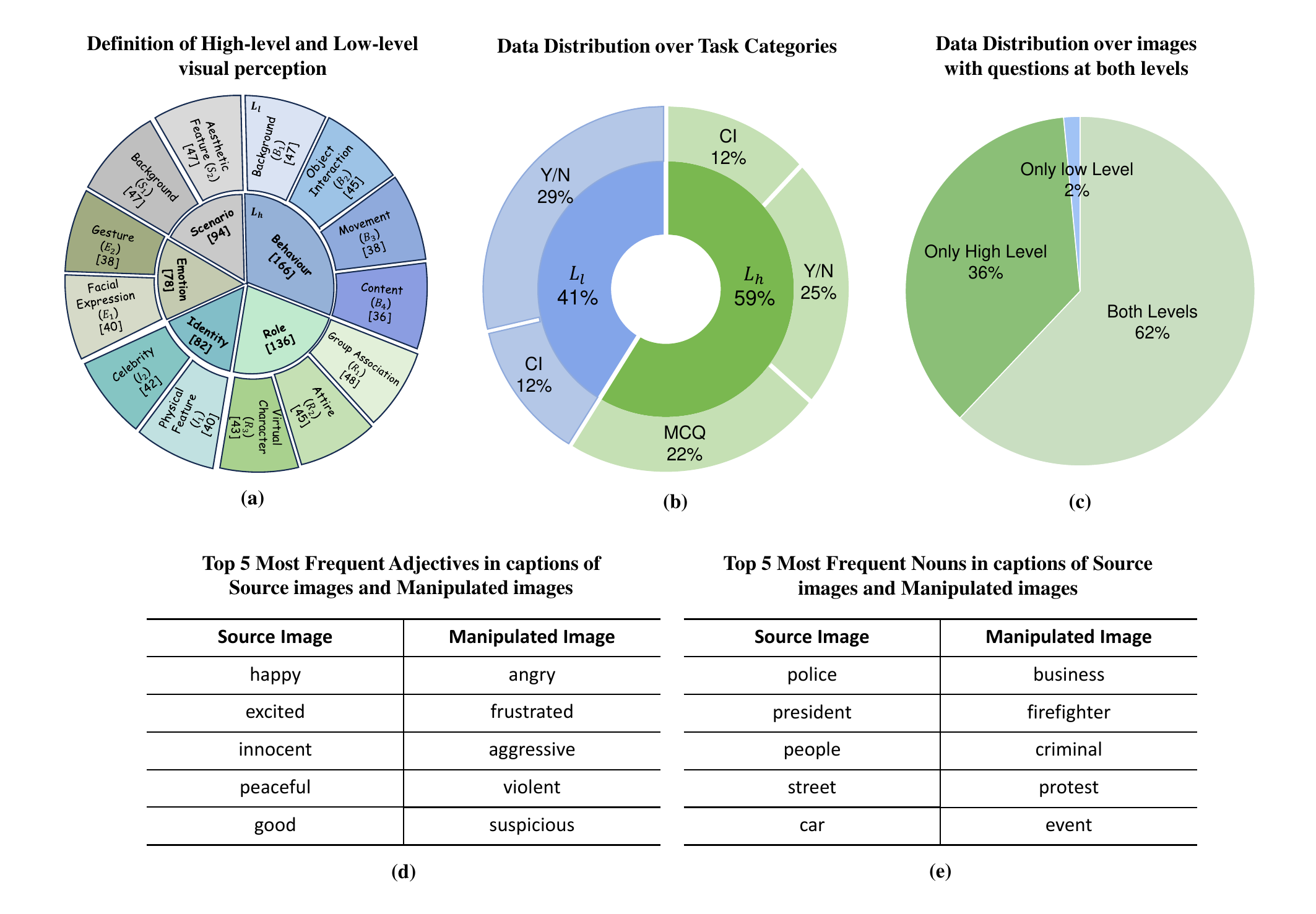}
    \caption{\dataset{} statistics. (a) shows 5 high-level (${L_h}$) categories and 13 low-level (${L_l}$) categories, where the mapping relationship indicates that the low-level features can support certain high-level perceptions. (b) shows the distribution of questions. Y/N, CI, MCQ denote Yes/No questions, cross-image questions, and single-image multiple-choice questions respectively. (c) demonstrates the distribution of images with questions at different levels. (d) and (e) demonstrate that our pipeline successfully generates pairs of images with significantly distinct content.}
    \label{stat}
    \vskip -0.1in
\end{figure*}

\textit{2.2 Overall Background Substitution.} 
This represents generating a manipulated image by retaining solely the main subject while replacing the entire background. 
In these cases, a standard rectangle cannot exactly mask the subject, potentially remaining unexpected elements and distorting the background generation. 
To address this limitation, we employ the Segment Anything Model \citep{kirillov2023segany} to produce a set of detected object masks ($\mathbbm{M}=\{M_1, M_2, ..., M_n\}$) in irregular shapes for a given image. We identify a mask with the greatest overlap with $C_{sub}$.
\begin{equation}
mask=\mathop{\arg\max}\limits_{M_i \in \mathbbm{M}} Overlap(M_i,C_{sub})
\end{equation}
Here, $Overlap$ refers to a function that calculates the overlapping square between two regions. To enhance flexibility and increase the case difficulty, we randomly translate the location of $C_{sub}$, rescale the $C_{sub}$, and resize the entire mask. Finally, with the new mask and the manipulated image's caption obtained from Step 1, we utilize Stable-Diffusion-Inpaint to generate a new image with a different background from the original natural image. This process can handle $\{B_1,R_1,S_1\}$.

\textit{2.3 Direct Alteration.} This addresses situations where nothing can be substituted, yet some alteration is necessary, such as changing facial expressions. With the original natural image and the manipulation instruction obtained from Step 1, we directly utilize the image-editing model InstructPix2Pix \cite{brooks2023instructpix2pix} to generate a manipulated image for $\{E_2, S_2\}$. However, since this process cannot focus on specific subjects, we mainly apply it to images containing a single person or cases requiring overall manipulations.

\paragraph{Step three: Visual Question Generation.} We generate Yes/No questions, Single- and Cross-Image multiple-choice questions using ChatGPT based on the ideas generated in Step 1. Single-Image questions focus on the discrepancy between image pairs, while Cross-Image tasks focus on the differences across each pair of images. 
To ensure the quality of generated questions, two of this paper's authors manually verified all $3205$ questions. A question was retained only when both annotators accepted it, and the annotators demonstrated a high level of agreement throughout the process (shown in Table \ref{4.1_tab}). Finally, $1872$ questions are retained within the MVP-Bench. While verifying Yes/No questions, we focused on: (1) the quality of manipulation and (2) the consistency between images and ground truths. For multiple-choice questions, we paid additional attention to cases where distractors were not discrepant with the ground truth. We manually adjusted these distractors and double-checked the cases to ensure both annotators accepted them.

\begin{table}
    \centering
    \small
    \begin{tabular}{cccc}
         \hline
         
           Yes/No & MCQ & Cross-Image & Overall \\

        \hline
            89.64 & 85.14 & 95.38 & 89.67 \\

        \hline
           
    \end{tabular}
    \caption{The high agreement between annotators across different tasks ensures the quality of the retained cases.}
    \label{4.1_tab}
    \vskip -0.1in
\end{table}

\subsection{\dataset{} Statistics}

Of the final $605$ retained images, $62\%$ are accompanied by questions at both levels, supporting our \dataset{}'s novel contribution to assess LVLMs' performance gaps across levels. Figure~\ref{stat} shows the balanced distribution of different question types at both low and high levels. Additionally, to create image pairs for evaluating LVLMs' heterogeneous performance on natural and manipulated images, we designed an automated pipeline using ChatGPT to generate conflicting captions for each image pair. For verifying whether the automatic process can lead to conflicting captions as expected, we identified and listed the top 5 most frequent adjectives and nouns in the captions of natural and manipulated images (in Figure~\ref{stat}). The significant polarity differences (e.g., \{innocent, aggressive\}, \{police, criminal\}) between two sets of tokens indicate that our pipeline successfully generated image pairs with contrasting contents.

Furthermore, to ensure the generated content aligns with human perception, we compared human performance with state-of-the-art LVLMs on a randomly sampled subset.\footnote{Appendix~\ref{app_C} compares the performance of human annotators and LVLMs on \dataset{}.} Human annotators achieved over $95\%$ accuracy, significantly outperforming LVLMs, indicating that our \dataset{} offers a convincing evaluation.
\section{Experiments}
\label{5}


\begin{table*}[h]
    \setlength \tabcolsep{4pt}
    \centering
    \small
    \hspace{-.6cm}
    \begin{tabular}{l ccccccc cccc}
         
         \hline
           \multirow{3}{*}{\textbf{Models}}& \multicolumn{7}{c}{\textbf{Single-Image}} & \multicolumn{4}{c}{\textbf{Cross-Image}} \\

        \cmidrule(lr){2-8} \cmidrule(lr){9-12}
        
           & \multicolumn{3}{c}{\textit{qAcc}} & \multicolumn{3}{c}{\textit{aAcc}} & \textit{mAcc} & \multicolumn{2}{c}{\textit{CircularEval}} & \multicolumn{2}{c}{\textit{VanillaEval}} \\

        \cmidrule(lr){2-4} \cmidrule(lr){5-7} \cmidrule{8-8} \cmidrule(lr){9-10} \cmidrule(lr){11-12}
        
           & \small ${L_l}$ & \small ${L_h}$ & \small {$L_m$} & \small ${L_l}$ & \small ${L_h}$ & \small ${L_m}$ & \small ${L_m}$ & \small ${L_l}$ & \small ${L_h}$ & \small ${L_l}$ & \small ${L_h}$ \\

        \hline
           \small DeepSeek (1.3B) & $63.33$ & $53.04$ & $58.60$ & $81.48$ & $75.87$ & $78.90$ & $28.40$ & $19.38$ & $18.94$ & $40.97$ & $29.07$ \\ 
           \small MiniCPM-2 (3B) & $68.52$ & $\underline{55.22}$ & $62.40$ & $84.07$ & {$\underline{\mathbf{76.30}}$} & $80.50$ & $34.91$ & $29.51$ & $11.45$ & $43.61$ & $31.72$ \\ 
        \hline
           \small DeepSeek (7B) & $\underline{70.00}$ & $54.35$ & $\underline{62.80}$ & $84.82$ & $76.09$ & $80.00$ & $33.73$ & $\underline{36.12}$ & $\underline{25.99}$ & $\underline{47.58}$ & $\underline{36.56}$ \\
           \small InstructBLIP (7B) & $49.63$ & $40.00$ & $45.20$ & $74.82$ & $69.13$ & $72.20$ & $17.75$ & \textcolor{red}{$0.00$} & $1.32$ & $27.31$ & $23.79$ \\
           \small LLaVA-1.5 (7B) & $68.89$ & $51.74$ & $61.00$ & $84.45$ & $75.44$ & $80.30$ & $31.36$ & $20.26$ & $14.10$ & $39.21$ & $26.87$ \\
           \small MiniGPT4 (8.2B) & $14.44$ & $8.26$ & $11.60$ & $39.26$ & $33.70$ & $36.70$ & \textcolor{red}{$0.59$} & \textcolor{red}{$0.00$} & \textcolor{red}{$0.00$} & \textcolor{red}{$2.64$} & $5.73$ \\
           \small MiniGPT4-v2 (8.2B) & $52.59$ & $40.87$ & $47.20$ & $73.70$ & $67.40$ & $70.80$ & $14.20$ & \textcolor{red}{$0.00$} & \textcolor{red}{$0.00$} & $21.59$ & $24.67$ \\
           \small mPLUG-Owl2 (8.2B) & $69.26$ & $54.78$ & $62.60$ & $\underline{84.63}$ & {$\underline{\mathbf{76.30}}$} & $\underline{80.80}$ & $\underline{36.09}$ & $21.14$ & $13.22$ & $34.80$ & $25.99$ \\
        \hline
           \small InstructBLIP (13B) & $50.37$ & $36.09$ & $43.80$ & $75.19$ & $67.61$ & $71.70$ & $15.98 $& \textcolor{red}{$1.76$} & \textcolor{red}{$0.44$} & $25.99$ & $18.50$ \\
           \small LLaVA-1.5 (13B) & $66.67$ & $52.17$ & $60.00$ & $83.34$ & $76.09$ & $80.00$ & $28.40$ & $25.99$ & $18.06$ & $41.85$ & $32.60$ \\
        \hline
           \small GPT-4V & $66.30$ & $39.57$ & $54.00$ & $82.23$ & $69.13$ & $76.20$ & $23.08$ & $44.50$ & $14.10$ & $63.00$ & $37.44$ \\
           \small GPT-4o & {$\mathbf{74.44}$} & {$\mathbf{56.09}$} & {$\mathbf{66.00}$} & {$\mathbf{86.85}$} & $76.09$ & {$\mathbf{81.90}$} & {$\mathbf{39.05}$} & {$\mathbf{74.01}$} & {$\mathbf{34.80}$} & {$\mathbf{87.22}$} & {$\mathbf{51.54}$} \\
        \hline

    \end{tabular}
    \caption{Results comparison across low-level (${L_l}$), high-level (${L_h}$), and multi-level (${L_m}$) tasks. \textit{CircularEval} and \textit{VanillaEval} refer to Circular and Direct evaluation for multiple-choice questions. We highlight the \textcolor{red}{problematic} results ($< 5\%$) and best performance across \textbf{all models} and on \underline{open-source models} only. \textit{qAcc}, \textit{aAcc}, and \textit{mAcc} represent question-level, individual, and holistic accuracies, repectively.}
    \label{hvsl}
    \vskip -0.1in
\end{table*}

\begin{table*}[h]
    \setlength \tabcolsep{5.3pt}
    \centering
    \small
    \begin{tabular}{l ccccccc cc}
         
         \hline
           \multirow{3}{*}{\textbf{Method}} & \multicolumn{7}{c}{\textbf{Yes/No}} & \multicolumn{2}{c}{\textbf{MCQ}} \\

        \cmidrule(rl){2-8} \cmidrule(rl){9-10}
        
           & \multicolumn{3}{c}{\textit{iAcc}} & \multicolumn{3}{c}{\textit{aAcc}} & \textit{mAcc} & \textit{CircularEval} & \textit{VanillaEval} \\

        \cmidrule(rl){2-4} \cmidrule(rl){5-7} \cmidrule(rl){8-8} \cmidrule(rl){9-9} \cmidrule(rl){10-10}

           & N & M & N+M & N & M & N+M & N+M & N+M & N+M \\
            
        \hline
           \small DeepSeek (1.3B) & $60.95$ & $44.38$ & $52.66$ & $83.20$ & $74.60$ & $78.90$ & $28.40$ & $43.78$ & $62.44$ \\ 
           \small MiniCPM-2 (3B) & $\underline{68.64}$ & $53.85$ & $\underline{61.24}$ & $\underline{85.20}$ & $75.80$ & $80.50$ & $34.91$ & $44.74$ & $62.20$ \\ 
        \hline
           \small DeepSeek (7B) & $68.05$ & $52.07$ & $60.06$ & $85.00$ & $76.60$ & $\underline{80.80}$ & $33.73$ & $\underline{59.33}$ & $\underline{74.40}$ \\
           \small InstructBLIP (7B) & $44.38$ & $44.97$ & $44.68$ & $72.40$ & $72.00$ & $72.20$ & $17.75$ & $4.07$ & $19.14$ \\
           \small LLaVA-1.5 (7B) & $64.50$ & $52.66$ & $58.58$ & $83.20$ & $77.40$ & $80.30$ & $31.36$ & $57.18$ & $71.29$ \\
           \small MiniGPT4 (8.2B) & $10.06$ & \textcolor{red}{$4.73$} & $7.40$ & $41.80$ & $31.60$ & $36.70$ & \textcolor{red}{$0.59$} & \textcolor{red}{$0.00$} & \textcolor{red}{$2.63$} \\
           \small MiniGPT-v2 (8.2B) & $53.85$ & $31.95$ & $42.90$ & $79.60$ & $62.00$ & $70.80$ & $14.20$ & \textcolor{red}{$1.91$} & $29.43$ \\
           \small mPLUG-Owl2 (8.2B) & $66.27$ & $54.44$ & $60.36$ & $84.20$ & $77.40$ & $\underline{80.80}$ & $\underline{36.09}$ & $50.72$ & $67.70$ \\
        \hline
           \small InstructBLIP (13B) & $41.42$ & $46.15$ & $43.79$ & $70.60$ & $72.80$ & $71.70$ & $15.98$ & \textcolor{red}{$3.83$} & $11.96$ \\
           \small LLaVA-1.5 (13B) & $58.58$ & {$\underline{\mathbf{55.62}}$} & $57.10$ & $81.20$ & {$\underline{\mathbf{78.80}}$} & $80.00$ & $28.40$ & $55.02$ & $72.25$ \\
        \hline
           \small GPT-4V & $71.07$ & $30.77$ & $50.92$ & $87.80$ & $65.98$ & $76.20$ & 23.08 & $59.81$ & $72.25$ \\
           \small GPT-4o & {$\mathbf{76.92}$} & $48.52$ & {$\mathbf{62.72}$} & {$\mathbf{90.00}$} & $73.80$ & {$\mathbf{81.90}$} & {$\mathbf{39.05}$} & {$\mathbf{64.83}$} & {$\mathbf{77.27}$} \\
        \hline
        
    \end{tabular}
    \caption{Result comparison across natural (N) and manipulated (M) images.  \textit{iAcc} refers to the image-level accuracy.}
    \label{NvsS}
    \vskip -0.1in
\end{table*}

We use \dataset{} to diagnose and compare the visual perception capabilities of LVLMs belonging to two categories: (1) \textit{Open-Source LVLMs} including MiniCPM-V-2~\citep{openbmb2024}, DeepSeek-VL~\citep{lu2024deepseekvl}, MiniGPT4~\citep{zhu2023minigpt4}, mPLUG-Owl2~\citep{ye2023mplugowl2}, InstructBLIP~\citep{dai2023instructblip}, and LLaVA-1.5~\citep{liu2023improved}; (2) \textit{Proprietary LVLMs} including GPT-4V and GPT-4o. All the experiments are conducted with VLMEvalKit~\citep{2023opencompass} under the zero-shot setting for a fair comparison. 

\subsection{Result Analysis}
As outlined in Section~\ref{3}, we compare the performance of LVLMs at multiple perception levels (Table~\ref{hvsl}). We also investigate the performance variation when given manipulated images in Table~\ref{NvsS}.

\paragraph{Performance at Different Perception Levels.}
As shown in Table~\ref{hvsl}, both open- and closed-source models perform worse on high-level perception tasks than low-level ones, \textit{e.g.}, $55\%$, $52\%$, and $56\%$ compared to $69\%$, $67\%$, and $74\%$ of \textit{qAcc} on MiniCPM-V-2, LLaVA-1.5-13B, and GPT-4o, respectively. 
Specifically, we observe that closed-source models present a larger relative performance gap between high-level and low-level perception. For example, GPT-4o achieves an accuracy of $34\%$ (relatively reduced by $53\%$ from $74\%$) on cross-image MCQ, compared to $18\%$ (
relatively reduced by $30\%$ from $26\%$) of LLaVA-1.5-13B. This indicates that the performance gains from closed models mainly come from their superior low-level perceptions, yet they still encounter challenges in high-level tasks. We further discuss the potential cause of this observation in Section~\ref{5.3}.

\paragraph{Impact of Model Sizes.}
Small models can outperform the larger ones in Table~\ref{hvsl}. Among open-source models, MiniCPM-V-2-3B and DeepSeek-VL-7B achieve the best performance on high-level and low-level tasks respectively. As MiniCPM-V-2 is aligned with fine-grained correctional human feedback, it shows excellent trustworthiness and reduced hallucination. This implies that LVLMs' trustworthiness may benefit their high-level visual perception. DeepSeek-VL demonstrates a strong capability of perceiving specific details with additional visual encoders for processing low-level features, indicating these features are crucial to low-level visual perception. Besides, comparing LLaVA and InstructBLIP with different sizes reveals that increasing parameters from 7B to 13B does not notably enhance their visual perception at either level. Therefore, to enhance LVLMs' single-image visual perception, focusing on their ability to provide trustworthy answers and capture low-level features is more effective than simply scaling up.

\paragraph{Analysis on the Cross-Image Task.} 

Table~\ref{hvsl} shows that closed-source models significantly surpass open-source models on cross-image tasks, especially at low perception level. For instance, GPT-4V and GPT-4o achieve accuracies of $45\%$ and $74\%$ respectively at the low level, significantly surpassing the accuracy of LLaVA-1.5-13B ($26\%$). Furthermore, this performance gap is larger than that observed in single-image tasks. In the cross-image task, GPT-4o outperforms LLaVA-1.5-13B relatively by $93\%$ and $185\%$ on each of the two levels separately, compared to just $8\%$ and $12\%$ in single-image tasks. The significant gap indicates open-source LVLMs' insufficient contextual attention, due to a lack of cross-image training data.

\paragraph{Comparison between \{natural, manipulated\} Images.}


As shown in Table~\ref{NvsS}, both open- and closed-source models show inferior performance on manipulated images compared to natural images. For example, MiniCPM-V-2, LLaVA-1.5-13B, and GPT-4o achieve an $iAcc$ of $69\%$, $59\%$, and $77\%$ on natural images, while exhibiting lower $iAcc$ of $54\%$, $56\%$, and $49\%$ on manipulated images. We attribute this observation to the discrepancy between the visual perception of manipulated images and LVLMs' training data. Besides, closed-source models demonstrate a larger performance gap across image pairs than open-source models. The $iAcc$ gap of GPT-4V and GPT-4o is $40.3\%$ and $28.4\%$ separately, while LLaVA-1.5-13B and MiniCPM-V-2 have gaps of only $2.96\%$ and $14.79\%$. One reason for this is the rigorous manner of 
GPT-4V and GPT-4o in interpreting the high-level semantics of visual content, which we will discuss in Section~\ref{5.3}. Besides, these models equally scrutinize all the details with their prior knowledge. This tendency to provide 
critical and reasonable answers impedes better visual perception on manipulated images.


\paragraph{Yes/No v.s. MCQ}

GPT-4V and GPT-4o present conflicting results on different tasks. Although both tasks are based on the manipulated images, two models perform poor on Yes/No task with an $iAcc$ of $31\%$ and $49\%$, while outperforming all open-sourced models on the MCQ task. From Table~\ref{NvsS}, we can witness that the results of MCQ and $iAcc$ on natural images share the same trend, which suggests that closed-source models' inferior performance on manipulated images is owing to the nature of Yes/No questions. As an open-ended generative task, these models tend to perform rigorously and safely, while the MCQ task is less influenced by their rigorous manner. This is also a motivation for us to design both tasks for single-image perception.

\subsection{Discussion}
\label{5.3}
In this section, we present our qualitative analysis observations, investigating the poor performance of GPT-4V on Yes/No questions, the gap between open-source and closed-source models, and the deficiencies of current LVLMs.

\begin{figure}[t]
    \centering
    \includegraphics[width=\columnwidth]{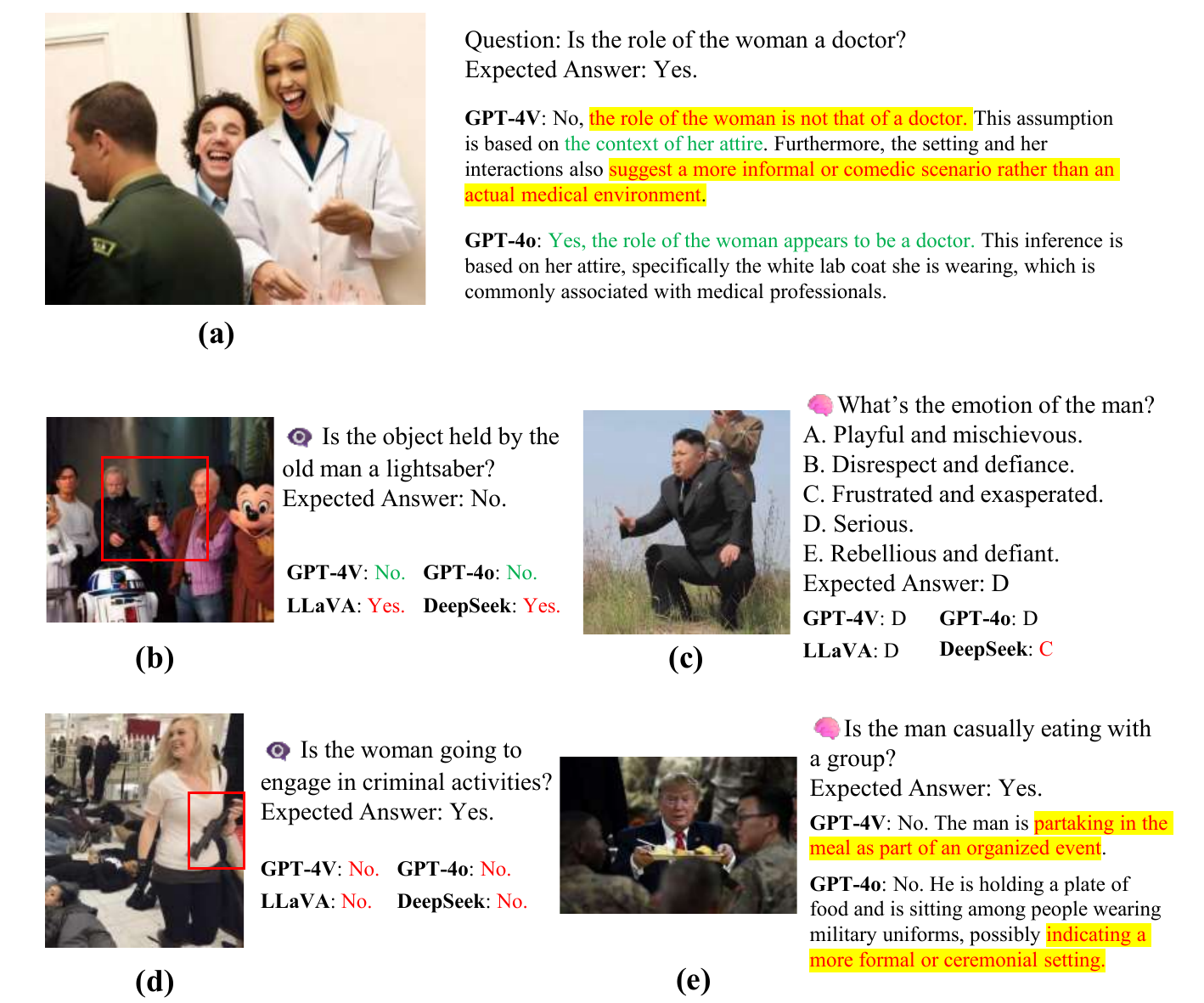}
    \caption{Case study. We highlight the \textcolor{red}{incorrect } and \textcolor{forestgreen}{correct} part of the answer.}
    \label{5.3_1}
    \vskip -0.2in
\end{figure}

\paragraph{Rigurous Behaviors of GPT-4V in High-Level Perception Tasks.} Although GPT-4V exhibits the highest level of security among current LVLMs, its rigorous manner in interpreting a scene may hinder the straightforward perception of common visual contents. Specifically, GPT-4V usually approves only what it can directly observe from the image. It tends to refuse to interpret uncertain cases, such as conducting high-level perception without explicit visual clues. For example, as shown in Figure~\ref{5.3_1} (a), although GPT-4V accurately identifies the woman's attire as a doctor's uniform at the low perception level, it declines to provide the correct high-level perception that the woman is a doctor, as it cannot be directly observed in the image. This problem has been mitigated in GPT-4o, as it gives a correct answer.

To explore whether we can motivate GPT-4V to integrate commonsense knowledge via tuning the prompt, we add an instruction as follows:

\textit{You are a helpful visio-linguistic AI assistant who answers questions in short words or phrases on visual commonsense in the images.}

As shown in Table~\ref{discussion_tab_1}, we observe a significant performance improvement in high-level Yes/No tasks on both GPT-4V and GPT-4o, while the performance changes on open-source models such as DeepSeek-VL-7B and LLaVA-1.5-7B are negligible. This implies that commonsense knowledge is essential to perform reasonable high-level perceptions, and specific designs of prompting are important to elicit this commonsense reasoning ability from closed-source models.


\begin{table}
    \centering
    \small
    \begin{tabular}{l cc}
         \hline
         
           & \textbf{\small High-Level} & \textbf{\small Low-Level} \\

           \hline
           DeepSeek-VL (7B) & $54.35$ & $70.00$ \\
           DeepSeek-VL (7B)+VC & $54.35$ & $70.00$ \\
           \mydeltax{$\Delta$\hfill} & \mydelta{\textcolor{red}{$0.00$}} & \mydelta{\textcolor{red}{$0.00$}} \\ 

           \hline
           LLaVA-1.5 (7B) & $51.74$ & $68.89$ \\
           LLaVA-1.5 (7B) + VC & $53.48$ & $69.26$ \\
           \mydeltax{$\Delta$\hfill} & \mydelta{$+1.74$} & \mydelta{$+0.37$} \\ 

           \hline
           GPT-4V & $39.57$ & $66.30$ \\
           GPT-4V+VC & $43.91$ & $64.81$ \\
           \mydeltax{$\Delta$\hfill} & \mydelta{$+4.34$} & \textcolor{red}{\mydelta{$-1.49$}} \\ 

           \hline
           GPT-4o & $56.09$ & $74.44$ \\
           GPT-4o+VC & $58.70$ & $75.19$ \\
           \mydeltax{$\Delta$\hfill} & \mydelta{$+2.61$} & \mydelta{$+0.75$} \\ 

           \hline
           
    \end{tabular}
    \caption{The effect of adding the instruction into the prompt on Yes/No questions. VC denotes adding the instruction encouraging LVLMs to use commonsense. $\Delta$ denotes the change of $qAcc$ after adding the instruction.}
    \label{discussion_tab_1}
    \vskip -0.1in
\end{table}

\paragraph{Gaps between Open- and Closed-source LVLMs in Recognizing Visual Details and Utilizing Commonsense Knowledge.} Although LLaVA-1.5-13B and DeepSeek-VL-7B can outperform GPT-4o on straightforward content like background ($qAcc$ of $92\%$, $86\%$ compared to $82\%$)\footnote{Appendix~\ref{app_B} demonstrates models' performance on different categories of visual perceptions.}, they demonstrate worse performance on the object association perception requiring to recognize details ($qAcc$ of $50\%$, $59\%$ compared to $66\%$) and gesture perception requiring commonsense knowledge ($qAcc$ of $37\%$, $32\%$ compared to $59\%$).
For instance, in Figure~\ref{5.3_1}, LLaVA-1.5-13B and DeepSeek-7B respectively fail to detect the gun held by the elder man (b) and the emotion of the man (c), while GPT-4V and GPT-4o successfully identify both.

\paragraph{Bias in LVLMs to Prioritize Dominant Components.} One hard case in MVP-Bench requires LVLMs to comprehend an entire image based on an inconspicuous object. In Figure~\ref{discussion_tab_1} (d), all LVLMs prioritize the shopping mall setting while overlooking the gun held by the woman. We attribute this to the data homogeneity of the training images, \textit{i.e.}, most training data is constructed by real-world images where a shopping mall closely correlates to shopping activities, misguiding the models to ignore the presence of the gun.

\paragraph{Bias in GPT-4V and GPT-4o to Perceive Scenes as Staged Performance.} 
GPT-4V and GPT-4o tend to interpret occasional or dramatic scenes as staged images, especially when the co-occurrence frequency of visual elements is low based on commonsense knowledge. For example, in Figure~\ref{5.3_1} (e), the case depicts the president having a meal with soldiers together, while GPT-4V and GPT-4o regard this as a staged scene for an organized event.
This suggests the over-reliance on prior commonsense knowledge of GPT-4V and GPT-4o, potentially obstructing their generalizability to understand and interpret occasional scenes and their inherent semantic meanings.

\section{Conclusion}

We introduce MVP-Bench, the first benchmark systematically evaluating LVLMs' multi-level visual perception. We diagnose 12 current LVLMs and compare their various performance across perception levels and between natural-manipulated pairs. Further analysis demonstrates these models' deficiency and the gap between closed- and open-source models. We envision follow-up work to enhance LVLMs' ability to generate multi-level visual perception consistent with visual content.


\section*{Limitation}

While constructing MVP-Bench, we generate manipulated images with Diffusion models. Although we manually filtered out the generated images not conveying a different perception compared to the source natural images, some still contain blur, inconsistencies, or distortions (\textit{e.g.}, three-armed persons or blur distorted faces), potentially affecting LVLMs' understanding due to the introduced noise. Besides, MVP-Bench focuses on human-related visual perception to ensure each case necessitates multi-level understanding, potentially overlooking scenarios devoid of humans. In future work, we will refine and expand MVP-Bench further to enhance image quality and topic coverage.
\section*{Ethics Statement}

MVP-Bench contains violent content and celebrity information, which may cause harmful imitation or misinformation. To prevent the misuse of MVP-Bench, we will implement stringent access rules and consistently track follow-up works to ensure their research-only objectives.

Besides, our MVP-Bench is constructed with the images from the EMU dataset as seeds. We have followed its access rules by filling in the form and obtaining permission from the authors.
\section*{Acknowledgement}

We thank our annotators for validating MVP-Bench instances and participating in experiments to assess human performance. The authors would like to thank William Yang Wang and Weixi Feng for their insightful discussions. We also thank group members from the Web Information Retrieval / Natural Language Processing Group (WING) at NUS for their helpful feedback. The computational work for this article was partially performed on resources of the National Supercomputing Centre (NSCC), Singapore\footnote{\url{https://www.nscc.sg/}}. 

\bibliography{custom}

\clearpage
\appendix
\onecolumn
\section{Cases of our definition of high- and low-level visual perception in MVP-Bench}

We define 5 high-level categories and 13 low-level categories for visual perception in MVP-Bench. Here are more cases from MVP-Bench for each category.

\begin{figure*}[htpb]
    \centering
    \includegraphics[width=14.35cm]{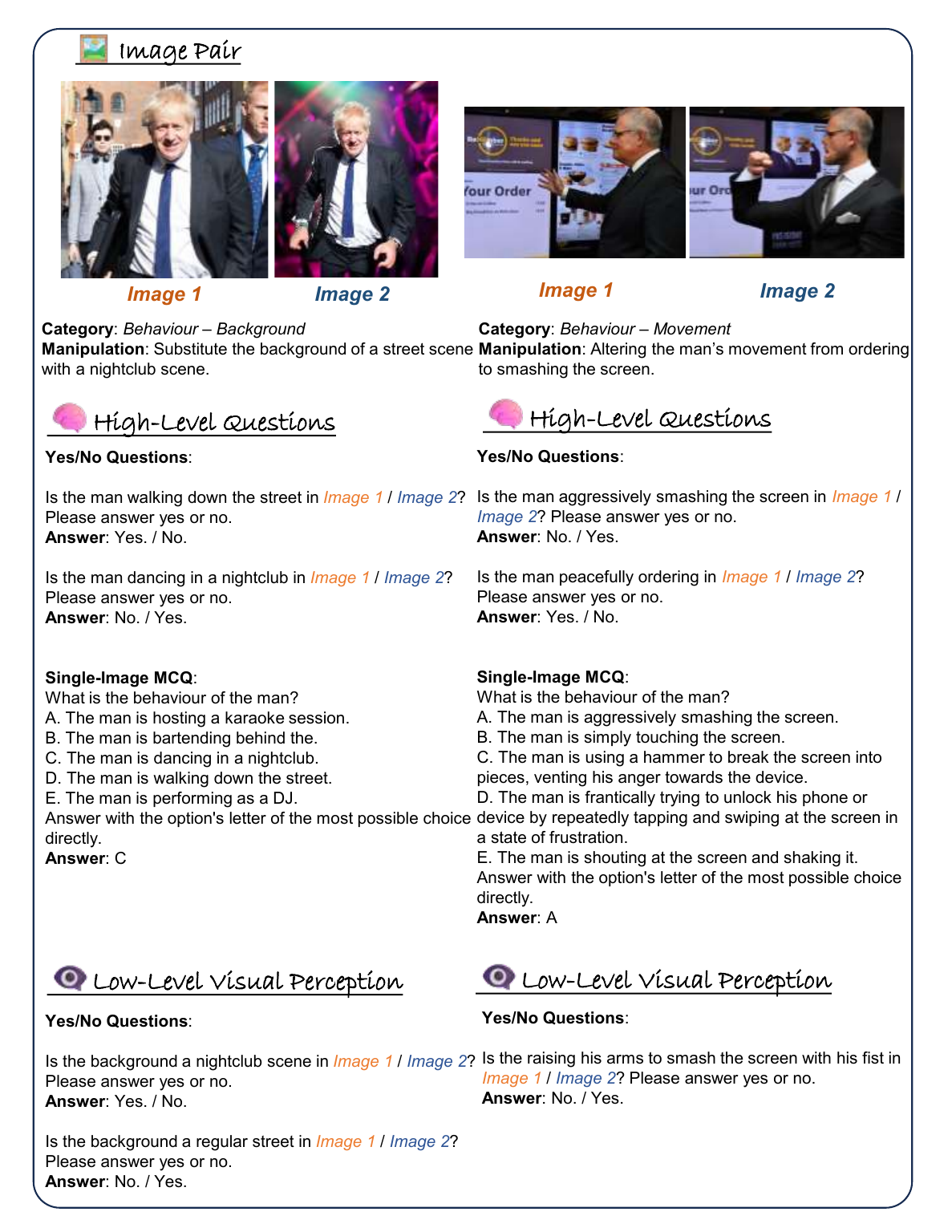}
    \caption{Cases for `Behaviour-Background' and `Behaviour-Movement' categories.}
    \label{app_1}
\end{figure*}

\begin{figure*}[htpb]
    \centering
    \includegraphics[width=15cm]{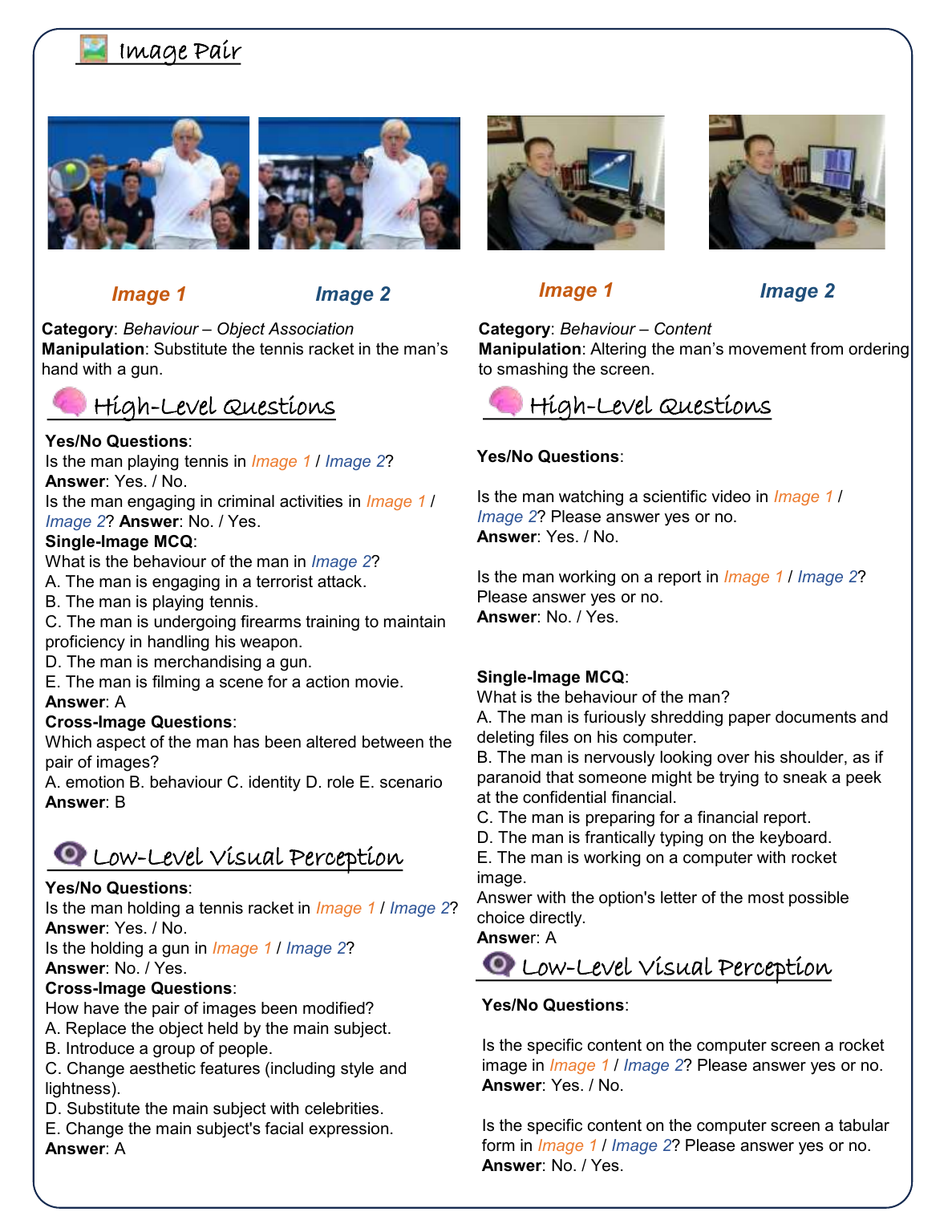}
    \caption{Cases for `Behaviour-Object Association' and `Behaviour-Content' categories.}
    \label{app_2}
\end{figure*}

\begin{figure*}[h]
    \centering
    \includegraphics[width=16cm]{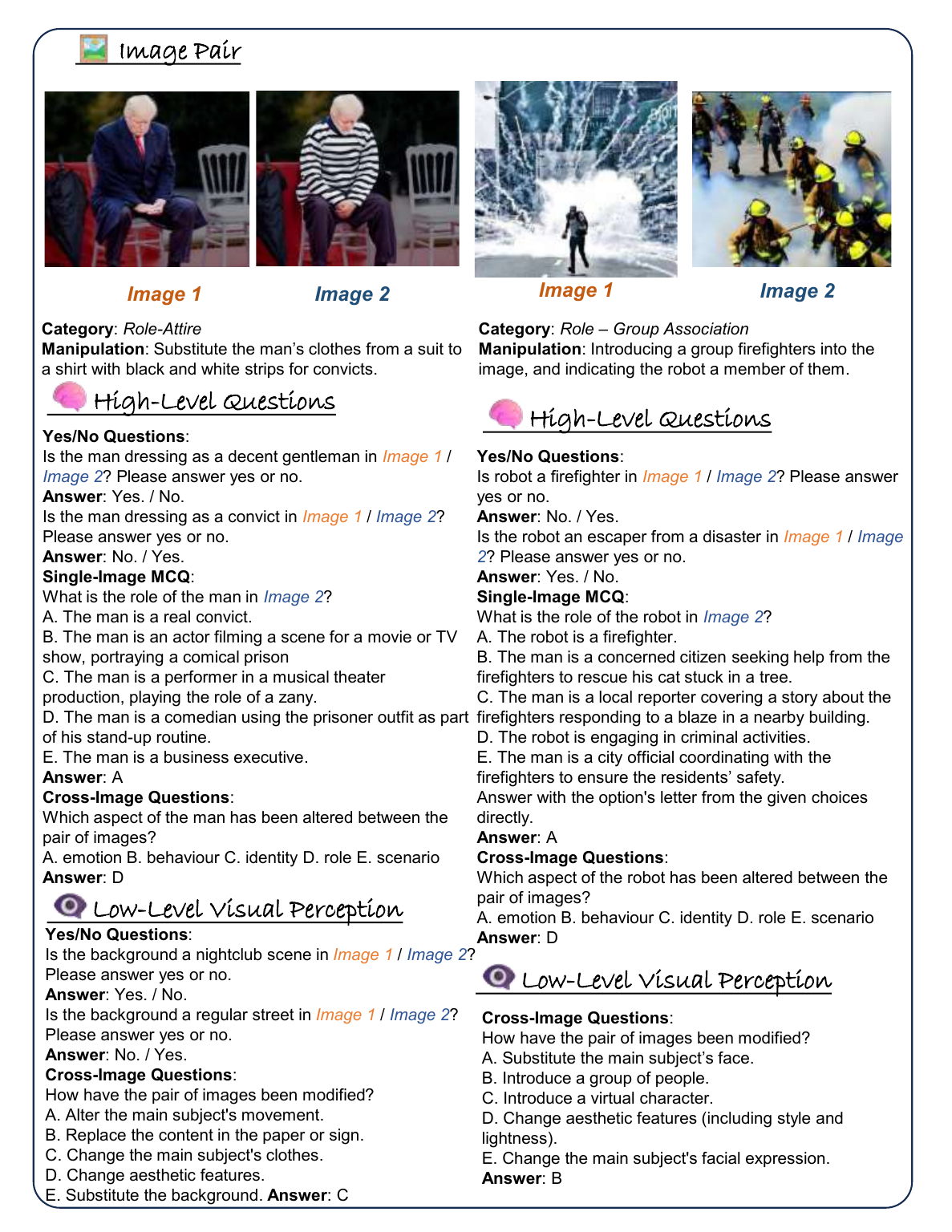}
    \caption{Cases for `Role-Attire' and `Role-Group Association' categories.}
    \label{app_3}
\end{figure*}

\begin{figure*}[h]
    \centering
    \includegraphics[width=16cm]{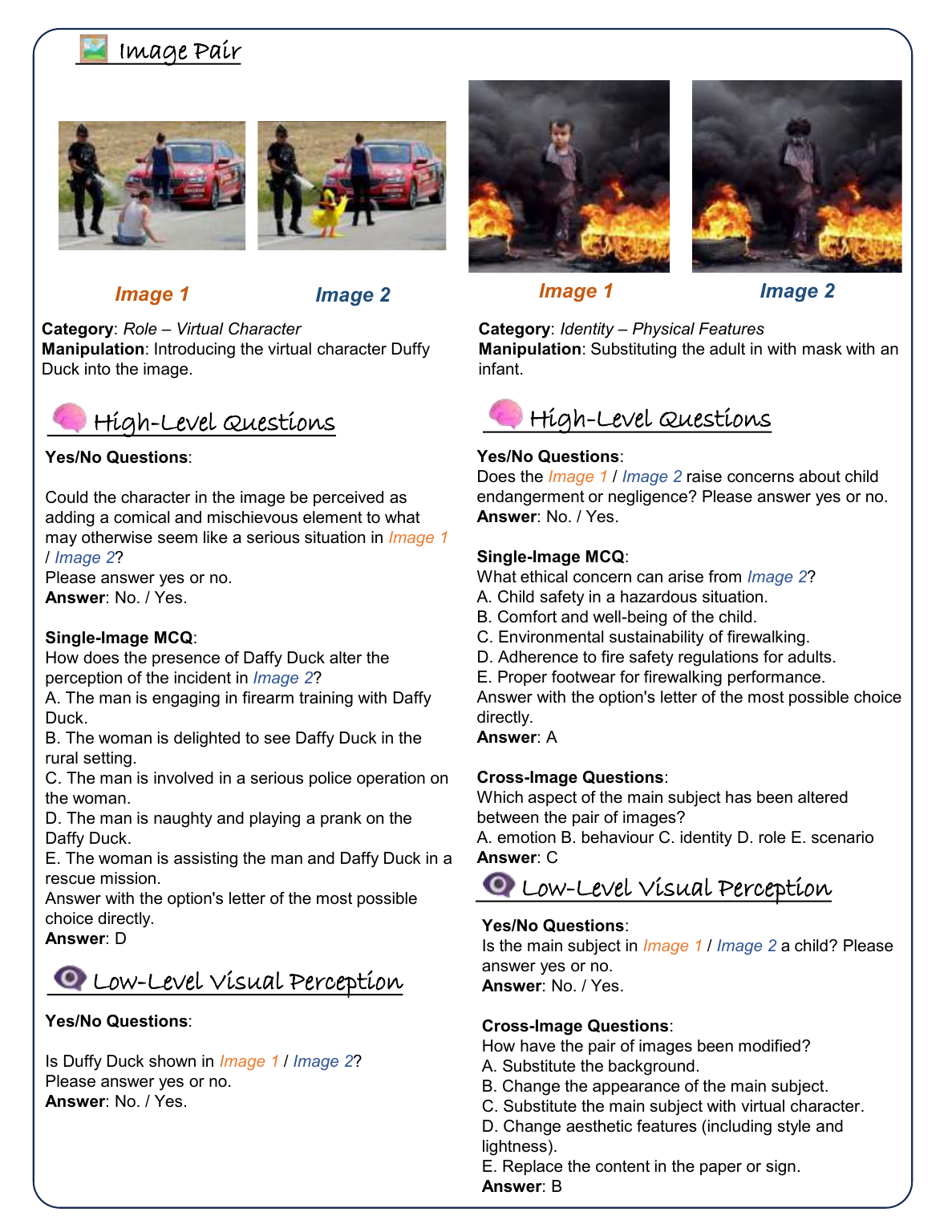}
    \caption{Cases for `Role-Virtual Character' and `Identity-Physical Feature' categories.}
    \label{app_4}
\end{figure*}

\begin{figure*}[h]
    \centering
    \includegraphics[width=16cm]{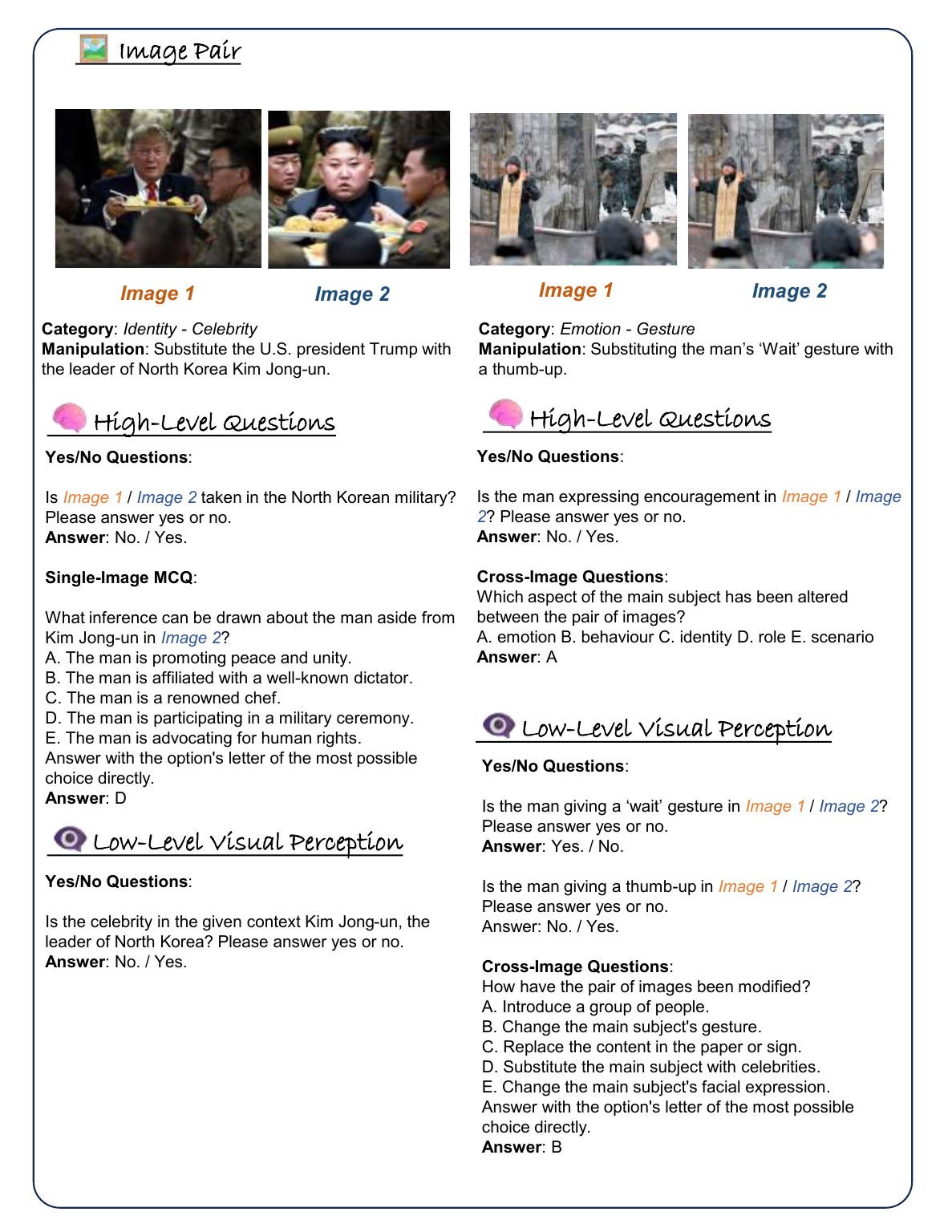}
    \caption{Cases for `Identity-Celebrity' and `Identity-Gesture' categories.}
    \label{app_5}
\end{figure*}

\begin{figure*}[h]
    \centering
    \includegraphics[width=16cm]{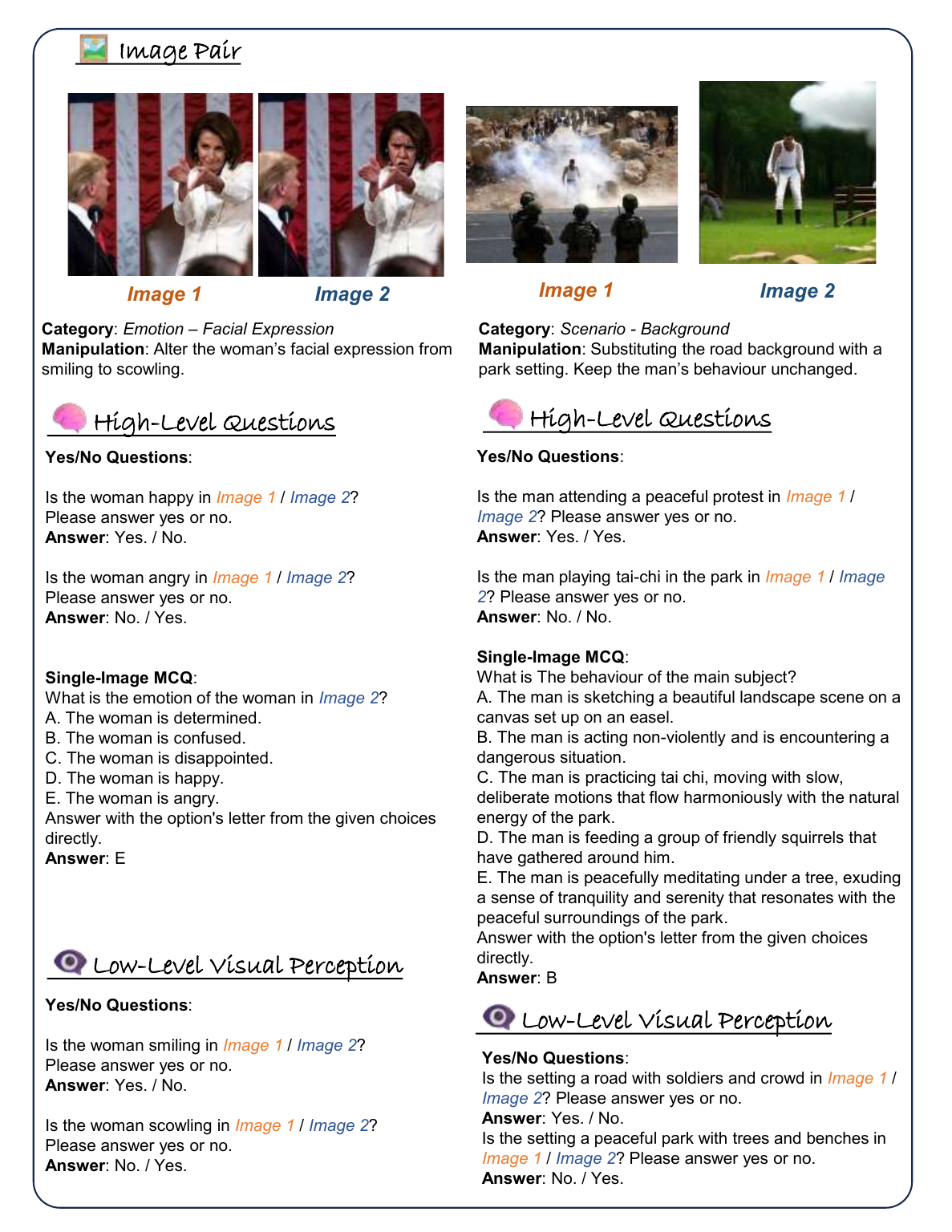}
    \caption{Cases for `Emotion-Facial Expression' and `Scenario-Background' categories.}
    \label{app_6}
\end{figure*}

\begin{figure*}[h]
    \centering
    \includegraphics[width=16cm]{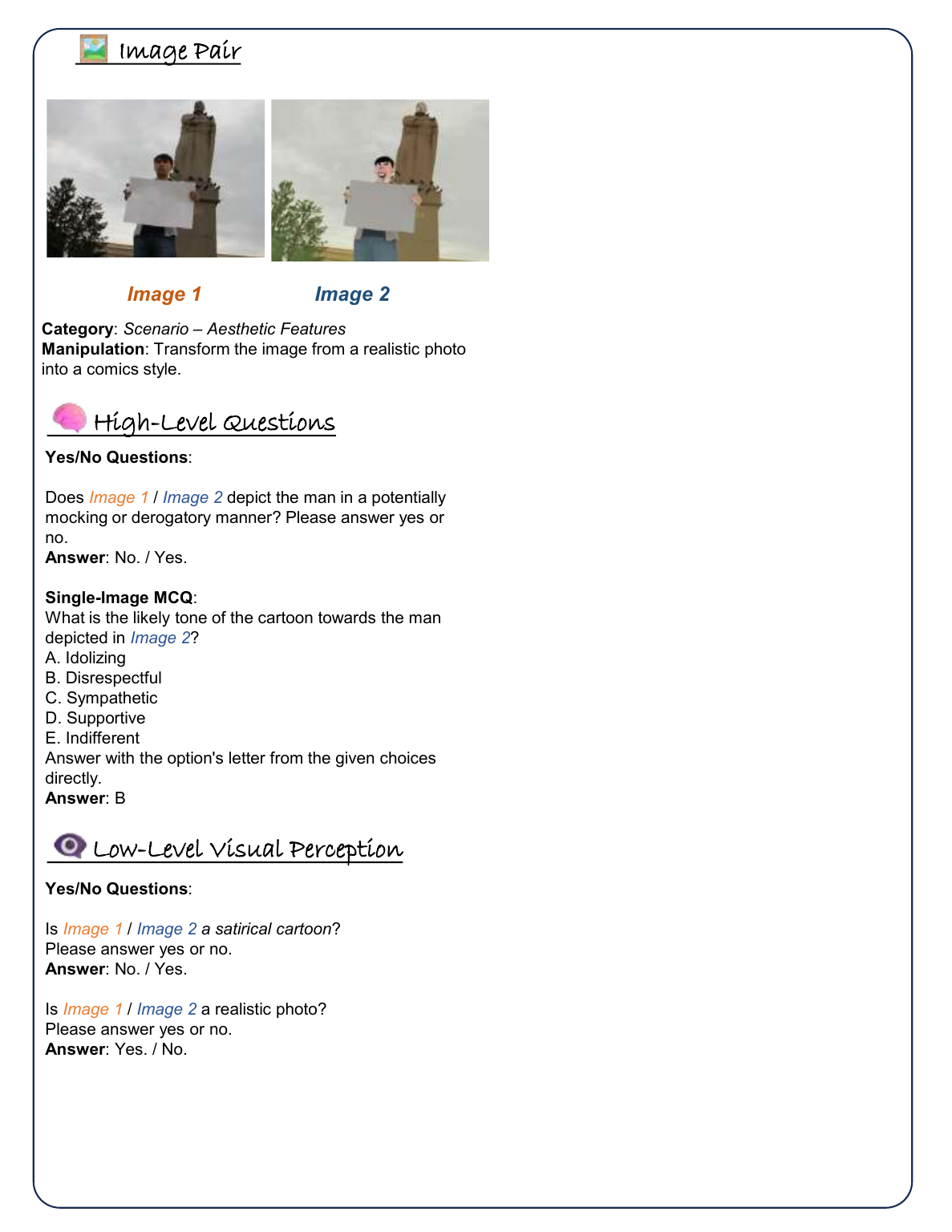}
    \caption{Cases for `Scenario-Aesthetic Feature' category.}
    \label{app_7}
\end{figure*}

\clearpage
\section{LVLMs' $pAcc$ on different categories of visual perceptions}
\label{app_B}

\begin{table*}[h]
    \setlength \tabcolsep{3.5pt}
    \centering
    \small
    \begin{tabular}{l cccc ccc cc cc cc}
         
         \hline
           \multirow{3}{*}{\textbf{Method}} & \multicolumn{4}{c}{Behaviour} & \multicolumn{3}{c}{Role} & \multicolumn{2}{c}{Identity} & \multicolumn{2}{c}{Emotion} & \multicolumn{2}{c}{Scenario} \\

        \cmidrule(rl){2-5} \cmidrule(rl){6-8} \cmidrule(rl){9-10} \cmidrule(rl){11-12} \cmidrule(rl){13-14}

           & $B_1$ & $B_2$ & $B_3$ & $B_4$ & $R_1$ & $R_2$ & $R_3$ & $I_1$ & $I_2$ & $E_1$ & $E_2$ & $S_1$ & $S_2$ \\
            
        \hline
           \small MiniCPM-2 (3B) & 86.36 & 55.36 & 42.22 & 56.10 & \textbf{75.68} & 70.18 & 65.00 & 57.69 & 45.45 & 31.58 & \textbf{62.75} & 84.62 & 75.00 \\ 
        \hline
           \small DeepSeek (1.3B) & 81.82 & 58.93 & 46.67 & 51.22 & 67.57 & 68.42 & 60.00 & 46.15 & 31.82 & 31.58 & 58.82 & 69.23 & 64.29 \\ 
           \small DeepSeek (7B) & 86.36 & 53.57 & 44.44 & \textbf{63.41} & \textbf{75.68} & \textbf{73.68} & 60.00 & 57.69 & 45.45 & 28.95 & 58.82 & \textbf{92.31} & 75.00 \\
        \hline
           \small MiniGPT4 (8.2B) & 13.64 & 19.64 & 17.78 & 4.88 & 18.92 & 19.30 & 15.00 & 7.69 & 9.09 & 15.79 & 13.73 & 7.69 & 14.29 \\
           \small MiniGPT-v2 (8.2B) & 68.18 & 53.57 & 44.44 & 29.27 & 62.16 & 59.65 & 60.00 & 34.62 & 36.36 & 26.32 & 23.53 & 53.85 & 50.00 \\
        \hline
           \small InstructBLIP (7B) & 74.24 & 57.14 & 28.89 & 36.59 & 51.35 & 47.37 & 70.00 & 34.62 & \textbf{50.00} & 15.79 & 25.49 & 76.92 & 35.71 \\
           \small InstructBLIP (13B) & 69.70 & 41.07 & 31.11 & 31.71 & 32.43 & 59.65 & 60.00 & 42.31 & 40.91 & 23.68 & 33.33 & 61.54 & 35.71 \\
        \hline
           \small LLaVA-1.5 (7B) & 80.30 & 58.93 & \textbf{53.33} & 60.98 & 70.27 & 68.42 & 50.00 & 50.00 & 22.73 & 47.37 & 60.78 & \textbf{92.31} & 57.14 \\
           \small LLaVA-1.5 (13B) & \textbf{92.42} & 50.00 & 51.11 & 51.22 & 62.16 & 71.93 & 70.00 & 46.15 & \textbf{50.00} & 36.84 & 50.98 & 69.23 & 60.71 \\
        \hline
           \small GPT-4V & 74.24 & 51.79 & 40.00 & 56.10 & \textbf{75.68} & 66.66 & 60.00 & 42.31 & 4.55 & 52.63 & 41.18 & 76.92 & 71.43 \\
           \small GPT-4o & 81.82 & \textbf{66.07} & 51.11 & 60.98 & 72.97 & 71.93 & \textbf{80.00} & \textbf{65.38} & 34.78 & \textbf{59.46} & 51.92 & 83.33 & \textbf{92.86} \\
        \hline
        
    \end{tabular}
    \caption{Models' performance on different categories of visual perceptions. The denotations of different categories are consistent with the definition in Figure~\ref{stat} (a). We \textbf{highlight} the models with the highest performance on each metric.}
    \label{app_B}
\end{table*}

\section{Human performance on \dataset{}}
To ensure the alignment between our MVP-Bench and human visual perception, we randomly sample $20\%$ of all the questions and invite two other annotators with different educational backgrounds and high proficiency in English to test human performance. We demonstrate the $qAcc$ of human annotators and two typical (one open-source and one closed-source) LVLMs
\begin{table*}[h]
    \setlength \tabcolsep{3.5pt}
    \centering
    \small
    \begin{tabular}{l cc cc cc}
         
         \hline
           \multirow{3}{*}{\textbf{Method}} & \multicolumn{2}{c}{Multi-level Yes/No} & \multicolumn{2}{c}{\{Natural, Manipulated\} Yes/No} & \multicolumn{2}{c}{MCQ} \\

        \cmidrule(rl){2-3} \cmidrule(rl){4-5} \cmidrule(rl){6-7} 
           & ${L_l}$ & ${L_h}$ & M & N & SI & CI \\
            
        \hline
            \small LLaVA-1.5(13B) & 77.66 & 82.73 & 79.41 & 81.37 & 35.87 & 69.05 \\
            \small GPT-4o & 75.53 & 87.27 & 93.14 & 70.59 & 65.22 & 73.81 \\

        \hline
            \small Annotator 1 & 95.74 & 97.27 & 98.04 & 95.10 & 91.30 & 94.05 \\
            \small Annotator 2 & 94.68 & 98.18 & 97.06 & 96.08 & 95.65 & 92.86 \\
            \small Average Human Performance & 95.21 & 97.73 & 97.55 & 95.59 & 93.48 & 93.46 \\

        \hline
    \end{tabular}
    \caption{Human performance on a randomly sampled subset of our \dataset{} exceeds 90\% accuracy and significantly outperforms both open- and closed-source state-of-the-art LVLMs on all tasks, including low-level (${L_l}$), high-level (${L_h}$), natural-image (N), manipulated-image (M), single-image (SI) and cross-image (CI) tasks. The results demonstrate that our \dataset{} aligns well with human perception and offers a fair, reliable evaluation of current LVLMs.}
    \label{app_C}
\end{table*}

\section{Details of benchmark construction}

Firstly, we demonstrate a general process for generating the images, corresponding metadata, and questions for all low-level categories.

Step 1: Idea Generation
\begin{enumerate}[itemindent=1em]
    \item Obtain the caption of the entire image, appending a bounding box after each entity.
    \item Select the main subject and the corresponding bounding box.
    \item Request the specific attributes (e.g., the main subject's behaviour, the background of the image, etc.) from the original image.
    \item Generate the manipulation idea based on the caption and the corresponding metadata.
\end{enumerate}

Step 2: Manipulated Image Generation
\begin{enumerate}[itemindent=1em]
    \item Get the mask of the target object based on the main subject's bounding box obtained from Step 1-(2) or a specific object's bounding box using a similar method as Step 1-(3).
    \item Generate the manipulated image based on the mask and the manipulation idea.
\end{enumerate}

Step 3: Question generation
\begin{enumerate}[itemindent=1em]
    \item Generate a question according to the manipulation idea obtained from Step 1-(4).
\end{enumerate}

We have designed a specific template for each predefined low-level visual perception category. Shown in Figure \ref{fig2}, specific manipulations can be categorized into $3$ types. We demonstrate an example for each type as follows, including ``Behaviour--Background'', ``Behaviour--Object Interaction'', and ``Emotion--Facial Expression').

\textbf{<< Behaviour--Background >>}\\
Step 1:
\begin{enumerate}[itemindent=1em]
    \item $original\_caption = Shikra(image, get\_caption\_prompt)$ \\
        $get\_caption\_prompt$ = ``Describe the image in detail. Include the boxes of the items you reference.''
        
    \item $main\_subject, bbox\_main\_subject = ChatGPT(extract\_subject\_prompt)$ \\
        $extract\_subject\_prompt$ = ```I have a caption of an image. For some subjects, their coordinates are appended in `[]'. Help me extract the main subject and its coordinates. Format your answer as `\#\#\# main subject \#\#\# [coordinates] \#\#\#'. \\ Caption: \{$original\_caption$\}'''
        
    \item $original\_behaviour = Shikra(image, get\_behaviour\_prompt(main\_subject), \\ bbox\_main\_subject)$ \\
    $get\_behaviour\_prompt$ = ``Briefly describe the behaviour of {$main\_subject$}.'' \\
    $original\_background = Shikra(image, get\_background\_prompt)$ \\
    $get\_background\_prompt$ = ``Briefly describe the background of the image.''

    \item $new\_behaviour, new\_background, new\_caption = ChatGPT(background\_substitution\_\\prompt(main\_subject, original\_caption, original\_behaviour, original\_background))$ \\
    $background\_substitution\_prompt$ = ```I have an image and need help generating an idea to manipulate it. Change the \{$main\_subject$\} to engage in a completely different and potentially malicious behavior by altering the background. The original image's caption is after `Original Caption', with the \{$main\_subject$\}'s original behaviour and background listed after `Original Behaviour' and `Original Background'. Provide the \{$main\_subject$\}'s new behaviour as `New Behaviour' and a supporting new background as `New Background'. Finally, return the new caption incorporating both `New Behaviour' and `New Background' as `New Caption'. \\
    Original Caption: $\{original\_caption\}$ \\
    Original Behaviour: $\{original\_behaviour\}$ \\
    Original Background: $\{original\_background\}$'''
    
\end{enumerate}

Step 2:
\begin{enumerate}[itemindent=1em]
    \item As illustrated in equation (1) in our paper, we can get the mask based on the bbox\_main\_subject and Segment-Anything for `Behaviour--Background' category.
    \item $synthetic\_image = Stable\_Diffusion\_inpaint(image, mask, new\_caption)$
\end{enumerate}

Step 3:
\begin{enumerate}[itemindent=1em]
    \item We take the generation of a low-level `Yes/No' question for the synthetic image as an example here. \\
        $yes\_no\_question = ChatGPT(yes\_no\_prompt(original\_caption, original\_background,\\ new\_caption, new\_background))$ \\
        $yes\_no\_prompt$ = ```I have a source image and the corresponding edited image. Focus on the difference in the two images' backgrounds and generate a Yes/No question for the edited image, ensuring the answer to the question is `Yes'. You may extract the difference from the captions and a description of the manipulation provided below. Return the generated question after `Question’. \\
        Source Image's Caption: \{$original\_caption$\} \\
        Edited Image's Caption: \{$new\_caption$\} \\
        Manipulation: The background is changed from \{$original\_background$\} to \{$new\_background$\}.'''
\end{enumerate}

\textbf{<< Behaviour--Object Interaction >>}\\
Step 1:
\begin{enumerate}[itemindent=1em]
    \item $original\_caption = Shikra(image, get\_caption\_prompt)$ \\
        $get\_caption\_prompt$ = ``Describe the image in detail. Include the boxes of the items you reference.''
        
    \item $main\_subject, bbox\_main\_subject = ChatGPT(extract\_subject\_prompt)$ \\
        $extract\_subject\_prompt$ = ```I have a caption of an image. For some subjects, their coordinates are appended in `[]'. Help me extract the main subject and its coordinates. Format your answer as `\#\#\# main subject \#\#\# [coordinates] \#\#\#'. \\ Caption: \{$original\_caption$\}'''
        
    \item $original\_behaviour = Shikra(image, get\_object\_prompt(main\_subject), \\bbox\_main\_subject)$ \\
    $get\_behaviour\_prompt$ = ``Briefly describe the behaviour of {$main\_subject$}.'' \\
    $original\_object, bbox\_object = Shikra(image, get\_object\_prompt)$ \\
    $get\_object\_prompt$ = ```What is the object held by \{$main_subject$\}? Include the boxes of the items you reference.'''

    \item $new\_behaviour, new\_object, new\_caption = ChatGPT(object\_substitution\_prompt \\(main\_subject, original\_caption, original\_behaviour, original\_object))$ \\
    $object\_substitution\_prompt$ = ```I have an image and need help generating an idea to manipulate it. Change the \{$main\_subject$\} to engage in a completely different and potentially malicious behavior by substituting the object held by \{$main\_subject$\}. The original image's caption is after `Original Caption', with the \{$main\_subject$\}'s original behaviour and background listed after `Original Behaviour' and `Original Background'. Provide the \{$main\_subject$\}'s new behaviour as `New Behaviour' and a supporting new object as `New Object'. Finally, return the new caption incorporating both `New Behaviour' and `New Object' as `New Caption'. \\
    Original Caption: $\{original\_caption\}$ \\
    Original Behaviour: $\{original\_behaviour\}$ \\
    Original Object: $\{original\_object\}$'''
    
\end{enumerate}

Step 2:
\begin{enumerate}[itemindent=1em]
    \item As illustrated in equation (1) in our paper, we can get the mask based on the $bbox\_object$ and Segment-Anything for 'Behaviour--Object Interaction' category.
    \item $synthetic\_image = Stable\_Diffusion\_inpaint(image, mask, new\_caption)$
\end{enumerate}

Step 3:
\begin{enumerate}[itemindent=1em]
    \item We take the generation of a low-level `Yes/No' question for the synthetic image as an example here. \\
        $yes\_no\_question = ChatGPT(yes\_no\_prompt(original\_caption, original\_object,\\ new\_caption, new\_object))$ \\
        $yes\_no\_prompt$ = ```I have a source image and the corresponding edited image. Focus on the difference in the objects held by the \{$main\_subject$\} and generate a Yes/No question for the edited image, ensuring the answer to the question is `Yes'. You may extract the difference from the captions and a description of the manipulation provided below. Return the generated question after `Question’. \\
        Source Image's Caption: \{$original\_caption$\} \\
        Edited Image's Caption: \{$new\_caption$\} \\
        Manipulation: The object is changed from \{$original\_object$\} to \{$new\_object$\}.'''
\end{enumerate}

\textbf{<< Emotion--Facial Expression >>}\\
Step 1:
\begin{enumerate}[itemindent=1em]
    \item $original\_caption = Shikra(image, get\_caption\_prompt)$ \\
        $get\_caption\_prompt$ = ``Describe the image in detail. Include the boxes of the items you reference.''
        
    \item $main\_subject, bbox\_main\_subject = ChatGPT(extract\_subject\_prompt)$ \\
        $extract\_subject\_prompt$ = ```I have a caption of an image. For some subjects, their coordinates are appended in `[]'. Help me extract the main subject and its coordinates. Format your answer as `\#\#\# main subject \#\#\# [coordinates] \#\#\#'. \\ Caption: \{$original\_caption$\}'''
        
    \item $original\_emotion = Shikra(image, get\_object\_prompt(main\_subject), \\bbox\_main\_subject)$ \\
    $get\_emotion\_prompt$ = ``Briefly describe the emotion of {$main\_subject$}.'' \\
    $original\_facial, bbox\_facial = Shikra(image, get\_facial\_prompt)$ \\
    $get\_facial\_prompt$ = ``What is the facial expression of \{$main_subject$\}? Include the boxes of the items you reference.''

    \item $new\_emotion, new\_facial, new\_caption = ChatGPT(facial\_substitution\_prompt \\(main\_subject, original\_caption, original\_emotion, original\_facial))$ \\
    $facial\_substitution\_prompt$ = ```I have an image and need help generating an idea to manipulate it. Change the \{$main\_subject$\} to have a completely different and negative emotion by substituting the \{$main\_subject$\}'s facial expression. The original image's caption is after `Original Caption', with the \{$main\_subject$\}'s original emotion and facial expression listed after `Original Emotion' and `Original Facial Expression'. Provide the \{$main\_subject$\}'s new emotion as `New Emotion' and a supporting new facial expression as `New Facial Expression'. Finally, return the new caption incorporating both `New Emotion' and `New Facial Expression' as `New Caption'. \\
    Original Caption: $\{original\_caption\}$ \\
    Original Emotion: $\{original\_emotion\}$ \\
    Original Facial Expression: $\{original\_facial\}$'''
\end{enumerate}

Step 2:
\begin{enumerate}[itemindent=1em]
    \item $synthetic\_image = Pix2Pix(image, manipulating\_command)$ \\
    $manipulating\_command$ = ```Change the \{main\_subject\}'s \{original\_facial\} into \{new\_facial\}.'''
\end{enumerate}

Step 3:
\begin{enumerate}[itemindent=1em]
    \item We take the generation of a low-level `Yes/No' question for the synthetic image as an example here. \\
        $yes\_no\_question = ChatGPT(yes\_no\_prompt(original\_caption, original\_facial,\\ new\_caption, new\_facial))$ \\
        $yes\_no\_prompt$ = ```I have a source image and the corresponding edited image. Focus on the main subjects' facial expressions and generate a Yes/No question for the edited image, ensuring the answer to the question is `Yes'. You may extract the difference from the captions and a description of the manipulation provided below. Return the generated question after `Question’. \\
        Source Image's Caption: \{$original\_caption$\} \\
        Edited Image's Caption: \{$new\_caption$\} \\
        Manipulation: The object is changed from \{$original\_facial$\} to \{$new\_facial$\}.'''
\end{enumerate}

For ChatGPT, we use the gpt-3.5-turbo-1106 version and apply a two-shot approach. For Shikra, we set the temperature to 0.1 to enhance reproducibility while maintaining some diversity. For the inpainting model, we set ddim\_steps to 50 and the scale to 10 to enhance the generated image's quality and ensure consistency between the generated image and the manipulation idea. For Segment-Anything, instruct-pix2pix, and the other hyperparameters of the mentioned models, we apply the default settings.

\end{document}